\documentclass[10pt,twocolumn,letterpaper]{article}

\usepackage{cvpr}
\usepackage{times}
\usepackage{epsfig}
\usepackage{graphicx}
\usepackage{amsmath}
\usepackage{amssymb}
\usepackage[accsupp]{axessibility}
\usepackage{algorithmic}
\usepackage{graphicx}
\usepackage{textcomp}
\usepackage{xcolor}
\usepackage{longtable}
\usepackage{subcaption}
 \usepackage{dblfloatfix}
\usepackage{float}
 \usepackage{booktabs}
\usepackage{bigstrut}
\usepackage{multirow}
\usepackage{color}
\usepackage{caption}
\usepackage{fancyhdr}
\usepackage{listings} %if lstlistings is used
\usepackage{changepage}
\usepackage{balance}
\usepackage{enumitem}
\usepackage{hyperref}
\usepackage{breqn}
\begin{document}

%%%%%%%%% TITLE
\title{Sound-Print: Generalised Face Presentation Attack Detection using Deep Representation of Sound Echoes
}
%Sound-Print: Generalised Face Presentation Attack Detection using Deep Representation of Time-Frequency features from Sound Echoes

\author{Raghavendra Ramachandra\textsuperscript{1} \quad \quad Jag Mohan Singh \textsuperscript{1}\quad \quad  Sushma Venkatesh\textsuperscript{2}   \\
\textsuperscript{1}Norwegian University of Science and Technology (NTNU), Norway.
\textsuperscript{2}AiBA AS, Norway. \\
%\textsuperscript{3}Indian Institute of Technology (IIT), Mandi.
%\textsuperscript{3}MOBAI AS, Norway.\\
{\tt\small email: \{raghavendra.ramachandra;jag.m.singh\}@ntnu.no; sushma@aiba.ai;}
}

%\and

\maketitle
\thispagestyle{empty}

%%%%%%%%% ABSTRACT
\begin{abstract}
Facial biometrics are widely deployed in smartphone-based applications because of their usability and increased verification accuracy in unconstrained scenarios. The evolving applications of smartphone-based facial recognition have also increased Presentation Attacks (PAs), where an attacker can present a Presentation Attack Instrument (PAI) to maliciously gain access to the application. Because the materials used to generate PAI are not deterministic, the detection of unknown presentation attacks is challenging. In this paper, we present an acoustic echo-based face Presentation Attack Detection (PAD) on a smartphone in which the PAs are detected based on the reflection profiles of the transmitted signal. We propose a novel transmission signal based on the wide pulse that allows us to model the background noise before transmitting the signal and increase the Signal-to-Noise Ratio (SNR). The received signal reflections were processed to remove background noise and accurately represent reflection characteristics. The reflection profiles of the bona fide and PAs are different owing to the different reflection characteristics of the human skin and artefact materials. Extensive experiments are presented using the newly collected Acoustic Sound Echo Dataset (ASED) with 4807 samples captured from bona fide and four different types of PAIs, including print (two types), display, and silicone face-mask attacks. The obtained results indicate the robustness of the proposed method for detecting unknown face presentation attacks.
 \end{abstract}

%%%%%%%%% BODY TEXT
\section{Introduction}
Face Recognition Systems (FRS) are vulnerable to presentation attacks that are  carried out by presenting facial artefacts to the face biometric capture system. The easy availability of the target face image that can be acquired by non-intrusive capture or through social networks makes these attacks common to operational FRS. Furthermore, Presentation Attacks (PAs) can easily be performed without any knowledge of the underlying operation of the biometric system. The wider deployment of the FRS in various applications, especially banking, has further elevated the PA on the FRS. 

\begin{figure}[t!]
\begin{center}
\includegraphics[width=1.0\linewidth]{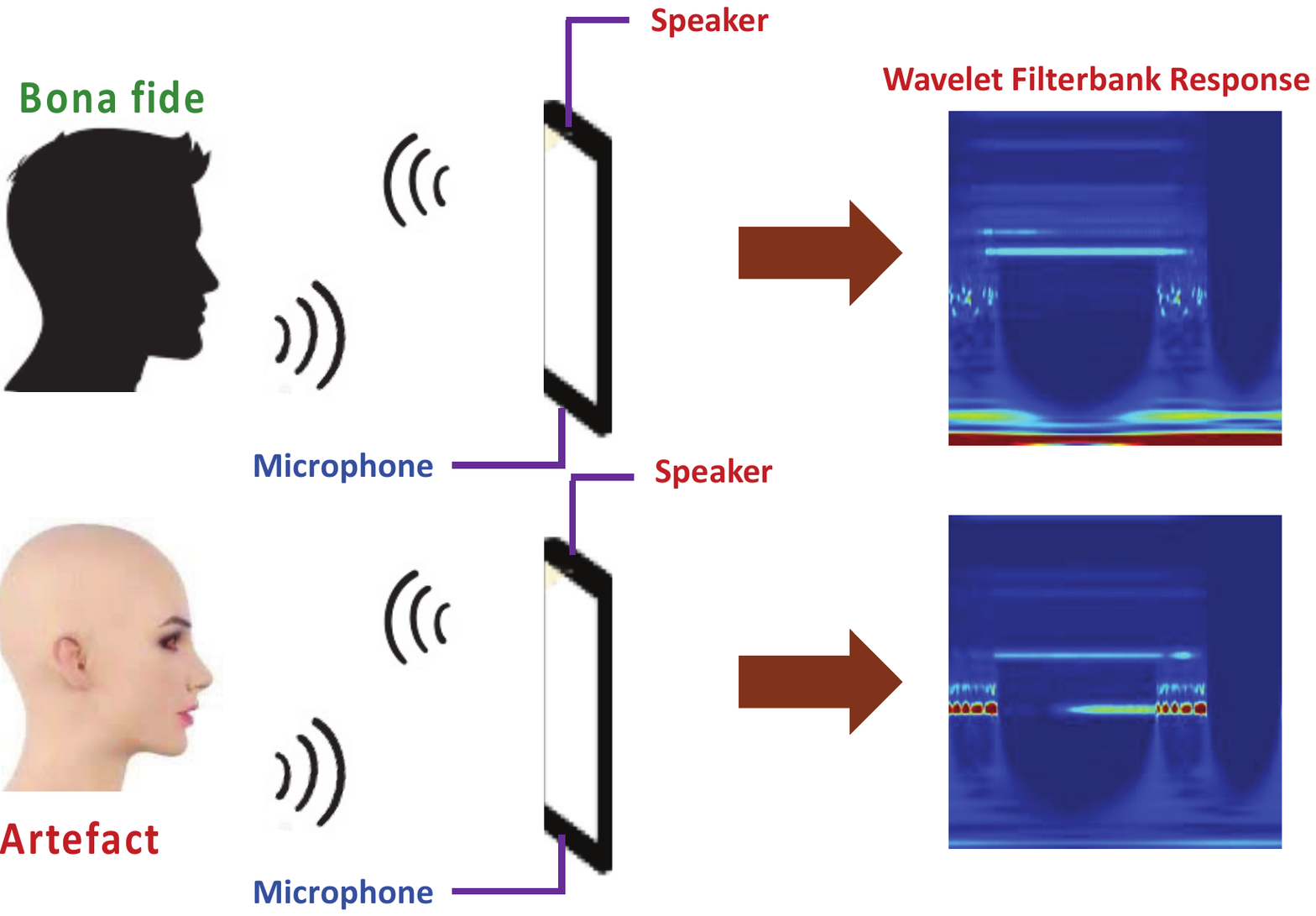}
\end{center}
   \caption{Illustration of Acoustic reflections for face presentation attack detection}
\label{fig:Intro}
\end{figure}
%--- Table SOTA %%%%%%
% \usepackage{graphicx}
\begin{table*}[htp]
\centering
\caption{Acoustics Face PAD State-Of-The-Art (SOTA) Techniques}
\resizebox{1\textwidth}{!}{%
\begin{tabular}{|l|l|l|l|l|l|}
\hline
\textbf{Authors} & \textbf{Transmission Signal} & \textbf{Types of PAIs} & \textbf{Database size} & \textbf{Multimodality} & \textbf{SmartPhone} \\ \hline
Zhou et al. \cite{zhou2018echoprint, zhou2021robust} & FWCM & \begin{tabular}[c]{@{}l@{}}Print Attack\\ Display Attack\end{tabular} & 45 Subjects & \begin{tabular}[c]{@{}l@{}}RGB face image\\ Acoustic reflections\end{tabular} & Samsung S7, S8 and Huawei P9 \\ \hline
Chen et al. \cite{chen2019echoface} & FWCM & Print Attack & 6 Subjects & Acoustic reflections & Samsung Galaxy C7 \\ \hline
Kong et al. \cite{kong2022beyond} & FWCM & \begin{tabular}[c]{@{}l@{}}Print Attack\\ Display Attack\end{tabular} & 30 Subjects & \begin{tabular}[c]{@{}l@{}}RGB face image\\ Acoustic reflections\end{tabular} & Samsung S9, S21, edge note and Xiaomi Redmi7 \\ \hline
\textbf{Our work} & \textbf{Wide Pulse} & \begin{tabular}[c]{@{}l@{}}\textbf{Print Attack}\\ \textbf{Silicone Mask}\\ \textbf{Display Attack}\end{tabular} & \textbf{35 Subjects} & \textbf{Acoustic reflections} & \textbf{Samsung S10} \\ \hline
\end{tabular}%
}
\label{tab:SOTA}%
\end{table*}
%--- END Table SOTA %%%%%%

The vulnerability of FRS to PAs has increased the interest of both academic and industrial researchers in developing Presentation Attack Detection (PAD) algorithms. Several high-end smartphones (e.g., Apple iPhone, Samsung) offer PAD by integrating multiple sensors (multispectral and 3D) and multibiometric systems. From an academic perspective, PAD algorithms have been extensively studied and broadly classified into hardware- and software-based methods \cite{yu2022deep, ramachandra2017presentation}. Hardware-based approaches employ additional hardware (such as multispectral cameras and liveness measuring devices) to reliably detect PAs at high cost and with limited scalability. Software-based approaches analyze captured face biometrics using either handcrafted features (using textures, gradients, and other image-based features) \cite{ramachandra2017presentation, heusch2020deep} or deep learning features \cite{yu2022deep, george2019deep, liu2021anomaly, yu2021revisiting, qin2021meta, yu2022benchmarking, wang2022patchnet, muhammad2022self, li2018unsupervised}. End-to-end deep learning techniques have been extensively studied in the literature on face PAD to achieve a reliable detection accuracy. However, generalizability across different types of PAIs for face PAD remains challenging.

In contrast to the existing techniques based on visual data (RGB images), acoustic echoes are used for face verification and PAs \cite{zhou2018echoprint}. Acoustic echo processing includes the sound signal transmitted using a smartphone speaker and recording sound reflections using microphones (see Figure \ref{fig:Intro}). %Because all existing smartphones have speakers and microphones, facial verification and PAD are scalable across many devices. 
The first work using acoustic echoes for both face verification and PAD was presented by Zhou et al. \cite{zhou2018echoprint, zhou2021robust}. A multimodal approach using acoustic and visual data was presented for verification and PAD. The underlying principle of using acoustic echoes for face verification assumes that the different 3D facial structure can reflect the sound signal with uneven attenuation at different frequencies that can be used for verification. For the PAD, the reflected signal has different reflection characteristics depending on the type of material used to generate the artefact. Therefore, the reflection profile was further processed to reliably detect the facial PAD.

In \cite{zhou2021robust}  Frequency-Modulated Continuous Wave (FMCW) signals with different frequencies were used as the transmitted signal, and the received signal was processed to recover echoes that were buried in the main lobes. Finally, a shallow serial Convolutional Neural Network (CNN) is proposed independently on acoustic echoes and facial images to perform verification and PAD. Chen et al. \cite{chen2019echoface} proposed the acoustic echo-based PAD in which the FMCW signal is used as the transmitted signal. The received signal was collected using two microphones located on the smartphones. In total, 16 chirp signals with different frequency ranges were used for transmission.   Recently, Kong et al. \cite{kong2022beyond}  presented a face PAD using acoustics and a face  (or RGB) image-based multimodal system. The transmitted signals were generated using the FMCW technique, with nine chirp signals at different frequencies. Detection was performed using a two-branch attention model trained separately on acoustic and RGB images. Based on the above discussion on SOTA methods, the FMCW signal is widely used to perform face verification and PAD. However, post-processing of the received echoes is challenging because they are buried in the strong main-lobe reflection and background noise. Furthermore, in previous approaches, experiments have been performed using only print and display attacks.

In this work, we present a novel method for face PAD based on acoustic signals on a smartphone. The main objective of the proposed method is to use only acoustic signals for the genralizable PAD.  The proposed method analyzes reflection echo characteristics to detect bona fides and PAs. The scattering property of the transmitted signal exhibits different characteristics owing to the change in the medium/material properties between the different types of PAIs and Bona fide. Therefore, the proposed method introduces a single wide pulse as the transmission signal to achieve a high Signal-to-Noise Ratio (SNR). Furthermore, the signal design also includes the silence period before transmission of the signal that will allow recording of the background noise, which is later subtracted from the received signal to reduce the background noise. After post-processing, the received signal is further represented by time-frequency components computed using Continuous Wavelet Transform (CWT) filter banks. The CWT representation was further processed through the pre-trained deep convolutional neural network (EfficientNet \cite{tan2019efficientnet}) to obtain deep features. Finally, PAD was performed using a linear SVM to effectively detect the PAs. Table \ref{tab:SOTA} summarizes the features of the proposed method and the existing SOTA method. Thus, the main contributions of this work are as follows: 

\begin{itemize} [leftmargin=*,noitemsep, topsep=0pt,parsep=0pt,partopsep=0pt]
\item We present a novel PAD based on acoustic signal processing for a face biometric system. We presented a wide pulse as the transmitted signal to achieve a high SNR. Furthermore, the proposed signal design incorporates a silent period to estimate the background noise. 
\item A new acoustic signal feature extraction method using CWT filter-banks and deep features are introduced to reliably detect the PAs. 
\item A new Acoustic Sound Echo Dataset (ASED) is collected with two different types of print PAIs, display attack PAI  and silicone face mask PAI, together with the bona fide samples from 35 unique data subjects resulting in 4807 acoustic samples.
\item Extensive experiments are performed to benchmark the generalizability of the proposed method across four different PAIs.  
\end{itemize}

\begin{figure*}[htp]
\begin{center}
\includegraphics[width=1.0\linewidth]{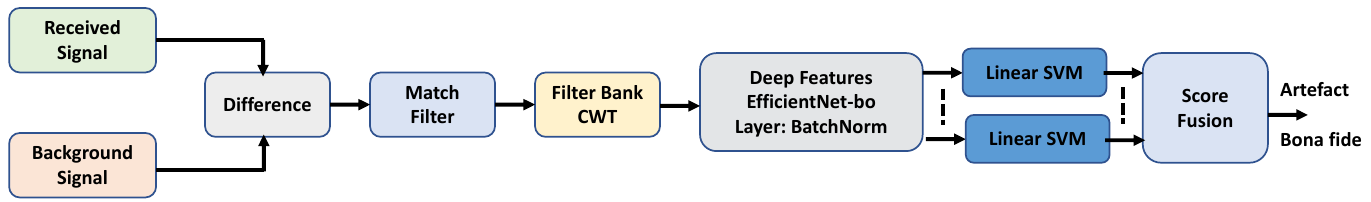}
\end{center}
   \caption{Block diagram of the proposed method}
\label{fig:Pro}
\end{figure*}

The remainder of the paper is organized as follows: Section \ref{sec:Prop} discusses the proposed method, Section \ref{sec:db} discusses the data collection protocols and dataset, Section \ref{sec:exp} presents the experimental results and discussion, and Section \ref{sec:Conc} concludes the paper. 
%------------------------------------------------------------------------
\section{Proposed Method}
\label{sec:Prop}

Figure \ref{fig:Pro} shows a block diagram of the proposed method that can be structured into six functional blocks: (1) transmitted signal design, (2) received signal processing, (3) filter bank using CWT, (4) deep features, (5) detector, and (6) fusion and final decisions. Each of these functional units of the proposed method is discussed as follows:  
\subsection{Transmitted sound signal design}
The primary goal of the proposed approach is to detect PAs based on the reflection characteristics recorded by the smartphone microphone. Therefore, designing a transmission signal is crucial for reliably characterizing reflections to detect PAs. In this work, we present a new signal considering two main fundamental characteristics: (a) selection of frequency and duration of the signal must enable reliable Signal-to-Noise Ratio (SNR) and (b) robustness to background noise.

The problem of detecting PA using  sound signals is based on analyzing the reflection characteristics of PAI materials. Therefore, the signal design must facilitate capturing the reflection from bona fide or PA artefacts and analyzing the reflection characteristics to detect PAs. Therefore, we are not interested in measuring the range (distance) and/or resolution (distinguishing between multiple objects) of the bona fide or the PA artefact. With this motivation, we introduce a wide rectangular pulse to achieve a sound beam (the straight line that this pulse can travel in space) that can sufficiently impact the bona fide face or PA artefact so that reflection with sufficient energy can be recorded to detect the PAs. Furthermore, the background noise must be effectively mitigated by sensing the environment to recover high-quality reflection. Therefore, the proposed signal design introduces a silent period before transmitting the pulse signal, during which the background noise is recorded. The recorded background signal can model the background while processing the received echoes to detect the PAs.

\begin{figure}[htp]
\begin{center}
\includegraphics[width=1.0\linewidth]{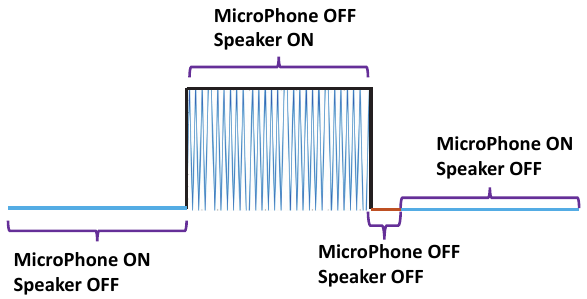}
\end{center}
   \caption{Proposed transmitted signal}
\label{fig:Signal}
\end{figure}

Figure \ref{fig:Signal} illustrates the shape of the transmitted sound signal introduced in this work. The process of sound signal transmission starts by setting the microphone ON and the speaker OFF. The microphone in the smartphone acts as the receiver, and the speaker is the transmitter that transmits the sound waves as shown in Figure \ref{fig:Signal}. The first part of the signal lasted for 1.5 seconds, during which the microphone was set ON to record the background noise. The  rectangular pulse is then transmitted through the speaker. The majority of existing smartphones support a 44.1 KHz sampling rate for microphones; therefore, the highest frequency sensed is 22kHz \cite{zhou2018echoprint}. Thus, the proposed signal adopts a rectangular pulse of 21KHz to sense the bona fide/PAs. The pulse duration lasted 2 seconds because the wider bandwidth permitted more energy to be recorded on the reflections received using the microphone.
Furthermore, the smartphone is held at a distance of 30 to 45 cm from the face during verification without obstacles, enabling richer and discriminant information from the reflected signal to detect the PAs reliably. After the sound signal is transmitted, both speaker and microphone are set to OFF for 0.5 seconds to avoid the chances of recording the transmitted signal. This step is essential to avoid direct interference, which can hide the reflection signals corresponding to the bona fide and/or PAs. In the last part of the signal, the microphone was set to ON, so that the reflections were recorded for 1.5 seconds. Thus, the total length of the signal is 1.5 (to record background) + 2 (transmitted signal) + 0.5 (idle time) + 1.5 (recording reflections) = 5.5 seconds.

\begin{figure*}[htp]
\begin{center}
\includegraphics[width=1.0\linewidth]{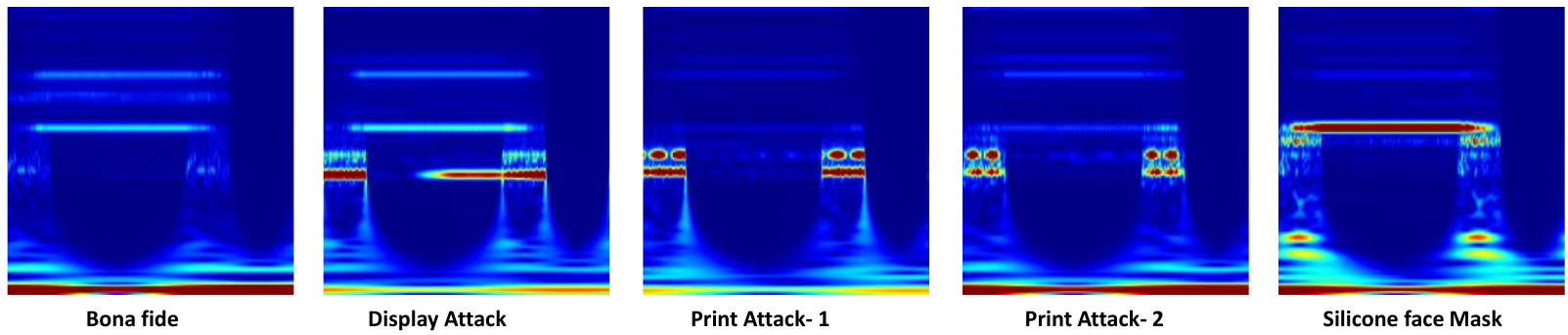}
\end{center}
   \caption{Qualitative results of CWT-FB for bona fide and PAs}
\label{fig:CWT}
\end{figure*}

\subsection{Received signal processing}
The received signal $R$ is first subtracted from the background signal $B$ to obtain clean signal $C_{s}$. In the next step, the clean signal $C_{s}$ is passed through a matched filter to maximize the signal-to-noise ratio, which is also known as pulse compression \cite{farnett1990pulse}. The match-filtering operation is performed by correlating the transmitted pulse with the received signal. The output of the matched filter $M_{c}$ is then used to detect the bona fide / PAs.

\subsection{Continuous Wavelet Transform Filter Bank (CWT-FB)}
In this work, we used the CWT-FB to extract the time-frequency information to capture the reflection characteristics of the bona fide and attack samples. Filter banks are designed for the length of the received signals and use a Morse Wavelet \cite{olhede2002generalized}. In this work, we choose a gamma value equal to 3 and a time-bandwidth product of 60 because, with gamma = 3, the Morse wavelet is perfectly symmetric in the frequency domain, allowing for a better capture of the time-frequency information. The designed CWT-FB comprises ten wavelet bandpass filters; therefore, the highest-frequency passband is set to 20Khz. Figure \ref{fig:CWT} illustrates the output of the CWT filter bank for both the bona fides and PAs employed in this study. These qualitative results indicate different time-frequency responses for the bona fide and PAs.

\subsection{Deep features}
In the next step, we extracted the features from the CWT-FB using off-the-shelf pre-trained Convolutional Neural Networks (CNN). In this work, we employed  EfficientNet b0 \cite{tan2019efficientnet} by considering the robustness and accuracy in several applications. Given the CWT-FB image, the deep features were extracted from the last Batch Normalization layer, resulting in $7$ $\times$ $7$ $\times$ $1280$ features. We chose the last BN layer based on an empirical analysis that indicated the best detection performance compared to the other layers in EfficientNet b0 \cite{tan2019efficientnet}.

\subsection{Detection Module}
The deep features are then used to train the classifiers to obtain the comparison scores. In this work, we employed  a linear Support Vector Machines (SVM) classifier to obtain the detection score. Since the deep features is of the dimension in $7$ $\times$ $7$ $\times$ $1280$, we employ  $7$ $\times$ $7$ = $49$ SVM classifiers that are trained independently on the $49$ different features of dimension $1$ $\times$ $1280$.

\subsection{Fusion and final decision}
Given the test vector of deep features of dimensions $7$ $\times$ $7$ $\times$ $1280$, we obtained 49 independent detection scores. Finally, the detection scores were fused using the sum rule to obtain the final detection score as follows: $D_{s} = \sum_{i = 1}^{49}  S_{i}$, where ${S_{i}}$ indicates the individual scores obtained using the SVM and $D_{s}$ indicates the final fused score. The final score $D_{s}$ is compared against the preset threshold to classify the received signal as bona fide / PA. 

%------------------------------------------------------------------------
\section{Acoustic Sound Echo Dataset (ASED)}
\label{sec:db}
In this work, we introduce a newly collected Acoustic Sound Echo Dataset (ASED) comprising 35 data subjects and four different PAIs, including two types of print attacks, display attacks, and silicone face masks. The proposed acoustic signaling (transmission and reflection) system was implemented as an Android application and was installed on a Samsung Galaxy S10. The data were collected in a laboratory setting, particularly in an indoor scenario reflecting the office environment. The user holds the phone so that the frontal camera can show the frontal face of the user. The angle of holding the phone is between 40-60 degrees such that the user can see the face image on the smartphone. Normally, the smartphone-to-face distance is between 20-40 cm. %Users can launch the application soon after seeing their face in the frontal camera. 
Bona fide data collection was conducted for 20 days in multiple sessions varying from 2 to 10 days, resulting in 35 to 40 samples for each data subject. We employed facial images from the data subjects to generate PAs using different types of  artefacts.

To capture the display attack, we employed iPad Pro 12.9, in which the face image was displayed on a smartphone to collect the data. To collect the print data, we used two different types of printers. Print-I: The data subject's face images were printed using a LaserJet printer with normal print paper. Print-II:  The data subject's face images are printed using the Dye Sublimation printer with a glossy paper. The use of two different types of printers allows for the analysis of the reflection characteristics of the proposed method for detecting PAs. The silicone face mask dataset was collected by wearing the silicone mask of the subject. Owing to the high cost of silicone masks, we used only four silicone face masks to collect the dataset. Thus, the ASED dataset comprised 1433 bona fides, 1234 display attacks, 500 print-I, 500 print-II, and 1140 silicone samples resulting in a total of . This resulted in 4807 samples, including bona fides and PAs.

\begin{table}[htp]
\centering
\caption{Statistics of Acoustic Sound Echo Dataset (ASED)}
\scriptsize
\resizebox{0.8\columnwidth}{!}{%
\begin{tabular}{@{}|l|l|l|@{}}
\toprule
Data Type       & Train Set & Test Set \\ \midrule \midrule
Bona fide       & 1003      & 430      \\ \midrule
Display Attack  & 899       & 385      \\ \midrule
Print-I Attack  & 350       & 150      \\ \midrule
Print-II Attack & 350       & 150      \\ \midrule
Silicone Attack & 798       & 342      \\ \bottomrule
\end{tabular}%
}
\label{tab:Stats}%
\end{table}
\subsection{Performance evaluation protocol:} 
We propose a protocol to evaluate the attack detection performance by dividing the entire dataset into two independent sets. The training set consisted of samples collected from 25 subjects and the testing set consisted of samples collected from 10 data subjects. Table \ref{tab:Stats} lists the statistics of the training and testing sample distributions used to evaluate PAD algorithms. However, for the silicone mask data, we have used two Silcone masks corresponding to unique identities for training and remaining  two for testing.
%------------------------------------------------------------------------
\section{Experiments and Results}
\label{sec:exp}

In this section, we present the quantitative performance of the proposed acoustic-based facial PAD technique. The performance of the face PAD was benchmarked using ISO/IEC  30107- 3 \cite{ISO-IEC-30107-3-PAD-metrics-170227} metrics such as Attack Presentation Classification Error Rate (APCER) and bona fide Presentation Classification Error Rate (BPCER). APCER is defined as the proportion of attack presentations incorrectly classified as bona fide, whereas BPCER is defined as the portion of the bona fide incorrectly classified as attack presentation. The Detection-Equal Error Rate (D-EER) indicates the value that the proportion of APCER is equal to the portion of BPCER. Extensive experiments were performed to benchmark the performance of the proposed method, highlighting the role of the background noise subtraction employed in the proposed transmission signal design. Furthermore, a comparison with the proposed feature extraction method using EfficientNet was benchmarked with other off-the-shelf pre-trained CNNs, such as DenseNet \cite{zhu2017densenet}, ResNet50 \cite{he2016deep} and MobileNetV2 \cite{sandler2018mobilenetv2}.

To effectively analyze the performance of the proposed method for generalizable PAD, we present quantitative results using two different protocols: inter and intra experiment. \textbf{Inter experiment protocol:} In this protocol, the PAD systems were trained and tested with different types of PAI. Hence, this protocol allowed us to analyze the generalizability of the proposed method to unknown PAI. \textbf{Intra experimental protocol:} In this protocol, the PAD system is trained and tested with the same type of PAI. Hence, this protocol allows the analysis of the robustness of the proposed method to known PAI.

%%%%%%%%%%%%%%%%%% without back ground Subtraction %%%%%%%%
\begin{table*}[!htb]
    \caption{Quantitative performance of the propose method with existing pre-trained \textbf{without back ground subtraction}}
    \begin{minipage}{.5\linewidth}
      %\caption{}
      \centering
\resizebox{.9\columnwidth}{!}{%
\begin{tabular}{|l|l|l|l|ll|}
\hline
\multirow{2}{*}{Algorithms} & \multirow{2}{*}{Train Data} & \multirow{2}{*}{Test Data} & \multirow{2}{*}{D-EER} & \multicolumn{2}{l|}{BPCER @ APCER =} \\ \cline{5-6} 
 &                           &          &       & \multicolumn{1}{l|}{5\%}   & 10\%  \\ \hline
\multirow{16}{*}{DenseNet}  & \multirow{4}{*}{Attack 1}   & Attack 1                   & 12.38                  & \multicolumn{1}{l|}{30.75}  & 14.31  \\ \cline{3-6} 
 &                           & Attack 2 & 12.16 & \multicolumn{1}{l|}{22.77} & 15.49 \\ \cline{3-6} 
 &                           & Attack 3 & 19.59 & \multicolumn{1}{l|}{55.16} & 27.94 \\ \cline{3-6} 
 &                           & Attack 4 & 12.15 & \multicolumn{1}{l|}{31.45} & 15.49 \\ \cline{2-6} 
 & \multirow{4}{*}{Attack 2} & Attack 1 & 11.27 & \multicolumn{1}{l|}{25.35} & 12.2  \\ \cline{3-6} 
 &                           & Attack 2 & 8.13  & \multicolumn{1}{l|}{11.13} & 7.98  \\ \cline{3-6} 
 &                           & Attack 3 & 10.18 & \multicolumn{1}{l|}{10.79} & 14.18 \\ \cline{3-6} 
 &                           & Attack 4 & 10.32 & \multicolumn{1}{l|}{16.43} & 10.79 \\ \cline{2-6} 
 & \multirow{4}{*}{Attack 3} & Attack 1 & 10.28 & \multicolumn{1}{l|}{24.41} & 10.32 \\ \cline{3-6} 
 &                           & Attack 2 & 14.65 & \multicolumn{1}{l|}{42.72} & 22.16 \\ \cline{3-6} 
 &                           & Attack 3 & 7.32  & \multicolumn{1}{l|}{12.22} & 5.86  \\ \cline{3-6} 
 &                           & Attack 4 & 9.79  & \multicolumn{1}{l|}{18.17} & 9.85  \\ \cline{2-6} 
 & \multirow{4}{*}{Attack 4} & Attack 1 & 6.81  & \multicolumn{1}{l|}{10.19} & 4.46  \\ \cline{3-6} 
 &                           & Attack 2 & 11.45 & \multicolumn{1}{l|}{22.77} & 14.78 \\ \cline{3-6} 
 &                           & Attack 3 & 8.71  & \multicolumn{1}{l|}{13.61} & 7.74  \\ \cline{3-6} 
 &                           & Attack 4 & 4.17  & \multicolumn{1}{l|}{3.99}  & 2.18  \\ \hline
\end{tabular}%
}
%\caption{}
    \end{minipage}%
    \begin{minipage}{.5\linewidth}
      \centering
\resizebox{.9\columnwidth}{!}{%
\begin{tabular}{|l|l|l|l|ll|}
\hline
\multirow{2}{*}{Algorithms} & \multirow{2}{*}{Train Data} & \multirow{2}{*}{Test Data} & \multirow{2}{*}{D-EER} & \multicolumn{2}{l|}{BPCER @ APCER =} \\ \cline{5-6} 
 &                           &          &       & \multicolumn{1}{l|}{5\%}   & 10\%  \\ \hline
\multirow{16}{*}{ResNet50}  & \multirow{4}{*}{Attack 1}   & Attack 1                   & 6.2                    & \multicolumn{1}{l|}{8.45}   & 4.69   \\ \cline{3-6} 
 &                           & Attack 2 & 4.69  & \multicolumn{1}{l|}{4.69}  & 3.28  \\ \cline{3-6} 
 &                           & Attack 3 & 12.14 & \multicolumn{1}{l|}{25}    & 15.49 \\ \cline{3-6} 
 &                           & Attack 4 & 4.97  & \multicolumn{1}{l|}{5.39}  & 1.46  \\ \cline{2-6} 
 & \multirow{4}{*}{Attack 2} & Attack 1 & 10.53 & \multicolumn{1}{l|}{19.15} & 11.5  \\ \cline{3-6} 
 &                           & Attack 2 & 7.44  & \multicolumn{1}{l|}{16.66} & 6.33  \\ \cline{3-6} 
 &                           & Attack 3 & 12.12 & \multicolumn{1}{l|}{27.23} & 14.55 \\ \cline{3-6} 
 &                           & Attack 4 & 9.14  & \multicolumn{1}{l|}{14.78} & 8.45  \\ \cline{2-6} 
 & \multirow{4}{*}{Attack 3} & Attack 1 & 14.12 & \multicolumn{1}{l|}{44.13} & 24.41 \\ \cline{3-6} 
 &                           & Attack 2 & 16.72 & \multicolumn{1}{l|}{51.64} & 32.62 \\ \cline{3-6} 
 &                           & Attack 3 & 8.14  & \multicolumn{1}{l|}{18.32} & 6.14  \\ \cline{3-6} 
 &                           & Attack 4 & 10.16 & \multicolumn{1}{l|}{26.76} & 10.32 \\ \cline{2-6} 
 & \multirow{4}{*}{Attack 4} & Attack 1 & 5.2   & \multicolumn{1}{l|}{7.51}  & 2.81  \\ \cline{3-6} 
 &                           & Attack 2 & 5.38  & \multicolumn{1}{l|}{8.45}  & 3.15  \\ \cline{3-6} 
 &                           & Attack 3 & 6.75  & \multicolumn{1}{l|}{7.74}  & 3.28  \\ \cline{3-6} 
 &                           & Attack 4 & 2.61  & \multicolumn{1}{l|}{1.87}  & 0.98  \\ \hline
\end{tabular}%
}
%\caption{}
    \end{minipage}

 \bigskip
 
    \begin{minipage}{.5\linewidth}
      %\caption{}
     \centering
\resizebox{0.9\columnwidth}{!}{%
\begin{tabular}{|l|l|l|l|ll|}
\hline
\multirow{2}{*}{Algorithms}   & \multirow{2}{*}{Train Data} & \multirow{2}{*}{Test Data} & \multirow{2}{*}{D-EER} & \multicolumn{2}{l|}{BPCER @ APCER =} \\ \cline{5-6} 
 &                           &          &       & \multicolumn{1}{l|}{5\%}   & 10\%  \\ \hline
\multirow{16}{*}{MobileNetV2} & \multirow{4}{*}{Attack 1}   & Attack 1                   & 10.77                  & \multicolumn{1}{l|}{19.48}  & 11.73  \\ \cline{3-6} 
 &                           & Attack 2 & 12.83 & \multicolumn{1}{l|}{31.22} & 17.61 \\ \cline{3-6} 
 &                           & Attack 3 & 13.51 & \multicolumn{1}{l|}{34.97} & 20.42 \\ \cline{3-6} 
 &                           & Attack 4 & 9.79  & \multicolumn{1}{l|}{15.72} & 9.35  \\ \cline{2-6} 
 & \multirow{4}{*}{Attack 2} & Attack 1 & 8.42  & \multicolumn{1}{l|}{11.97} & 8.21  \\ \cline{3-6} 
 &                           & Attack 2 & 8.47  & \multicolumn{1}{l|}{8.45}  & 7.51  \\ \cline{3-6} 
 &                           & Attack 3 & 8.7   & \multicolumn{1}{l|}{12.21} & 8.45  \\ \cline{3-6} 
 &                           & Attack 4 & 8.23  & \multicolumn{1}{l|}{8.45}  & 7.51  \\ \cline{2-6} 
 & \multirow{4}{*}{Attack 3} & Attack 1 & 10.77 & \multicolumn{1}{l|}{10.17} & 11.51 \\ \cline{3-6} 
 &                           & Attack 2 & 12.41 & \multicolumn{1}{l|}{23.23} & 13.84 \\ \cline{3-6} 
 &                           & Attack 3 & 9.27  & \multicolumn{1}{l|}{15.72} & 8.45  \\ \cline{3-6} 
 &                           & Attack 4 & 10.85 & \multicolumn{1}{l|}{14.78} & 11.26 \\ \cline{2-6} 
 & \multirow{4}{*}{Attack 4} & Attack 1 & 10.28 & \multicolumn{1}{l|}{15.72} & 10.32 \\ \cline{3-6} 
 &                           & Attack 2 & 13.97 & \multicolumn{1}{l|}{24.41} & 19.95 \\ \cline{3-6} 
 &                           & Attack 3 & 10.88 & \multicolumn{1}{l|}{35.44} & 11.73 \\ \cline{3-6} 
 &                           & Attack 4 & 7.32  & \multicolumn{1}{l|}{10.56} & 5.39  \\ \hline
\end{tabular}%
}
    \end{minipage}%
    \begin{minipage}{.5\linewidth}
     \centering
\resizebox{0.9\columnwidth}{!}{%
\begin{tabular}{|l|l|l|l|ll|}
\hline
 &
   &
   &
   &
  \multicolumn{2}{l|}{BPCER @ APCER =} \\ \cline{5-6} 
\multirow{-2}{*}{Algorithms} &
  \multirow{-2}{*}{Train Data} &
  \multirow{-2}{*}{Test Data} &
  \multirow{-2}{*}{D-EER} &
  \multicolumn{1}{l|}{5\%} &
  10\% \\ \hline
 &
   &
  Attack 1 &
  {\color[HTML]{3166FF} \textbf{3.46}} &
  \multicolumn{1}{l|}{{\color[HTML]{3166FF} \textbf{2.58}}} &
  {\color[HTML]{3166FF} \textbf{1.17}} \\ \cline{3-6} 
 &
   &
  Attack 2 &
  {\color[HTML]{3166FF} \textbf{4.29}} &
  \multicolumn{1}{l|}{{\color[HTML]{3166FF} \textbf{3.99}}} &
  {\color[HTML]{3166FF} \textbf{1.64}} \\ \cline{3-6} 
 &
   &
  Attack 3 &
  {\color[HTML]{3166FF} \textbf{6.75}} &
  \multicolumn{1}{l|}{{\color[HTML]{3166FF} \textbf{7.51}}} &
  {\color[HTML]{3166FF} \textbf{3.99}} \\ \cline{3-6} 
 &
  \multirow{-4}{*}{Attack 1} &
  Attack 4 &
  {\color[HTML]{3166FF} \textbf{2.61}} &
  \multicolumn{1}{l|}{{\color[HTML]{3166FF} \textbf{2.11}}} &
  {\color[HTML]{3166FF} \textbf{0.7}} \\ \cline{2-6} 
 &
   &
  Attack 1 &
  {\color[HTML]{3166FF} \textbf{9.41}} &
  \multicolumn{1}{l|}{{\color[HTML]{3166FF} \textbf{9.62}}} &
  {\color[HTML]{3166FF} \textbf{9.15}} \\ \cline{3-6} 
 &
   &
  Attack 2 &
  {\color[HTML]{3166FF} \textbf{8.58}} &
  \multicolumn{1}{l|}{{\color[HTML]{3166FF} \textbf{8.68}}} &
  {\color[HTML]{3166FF} \textbf{8.45}} \\ \cline{3-6} 
 &
   &
  Attack 3 &
  {\color[HTML]{3166FF} \textbf{9.51}} &
  \multicolumn{1}{l|}{{\color[HTML]{3166FF} \textbf{9.62}}} &
  {\color[HTML]{3166FF} \textbf{9.62}} \\ \cline{3-6} 
 &
  \multirow{-4}{*}{Attack 2} &
  Attack 4 &
  {\color[HTML]{3166FF} \textbf{8.88}} &
  \multicolumn{1}{l|}{{\color[HTML]{3166FF} \textbf{9.15}}} &
  {\color[HTML]{3166FF} \textbf{8.58}} \\ \cline{2-6} 
 &
   &
  Attack 1 &
  {\color[HTML]{3166FF} \textbf{3.46}} &
  \multicolumn{1}{l|}{{\color[HTML]{3166FF} \textbf{2.81}}} &
  {\color[HTML]{3166FF} \textbf{1.42}} \\ \cline{3-6} 
 &
   &
  Attack 2 &
  {\color[HTML]{3166FF} \textbf{1.37}} &
  \multicolumn{1}{l|}{{\color[HTML]{3166FF} \textbf{0.23}}} &
  {\color[HTML]{3166FF} \textbf{0.23}} \\ \cline{3-6} 
 &
   &
  Attack 3 &
  {\color[HTML]{3166FF} \textbf{1.49}} &
  \multicolumn{1}{l|}{{\color[HTML]{3166FF} \textbf{0.71}}} &
  {\color[HTML]{3166FF} \textbf{0.46}} \\ \cline{3-6} 
 &
  \multirow{-4}{*}{Attack 3} &
  Attack 4 &
  {\color[HTML]{3166FF} \textbf{1.82}} &
  \multicolumn{1}{l|}{{\color[HTML]{3166FF} \textbf{1.4}}} &
  {\color[HTML]{3166FF} \textbf{0.46}} \\ \cline{2-6} 
 &
   &
  Attack 1 &
  {\color[HTML]{3166FF} \textbf{9.41}} &
  \multicolumn{1}{l|}{{\color[HTML]{3166FF} \textbf{9.38}}} &
  {\color[HTML]{3166FF} \textbf{9.38}} \\ \cline{3-6} 
 &
   &
  Attack 2 &
  {\color[HTML]{3166FF} \textbf{9.39}} &
  \multicolumn{1}{l|}{{\color[HTML]{3166FF} \textbf{9.38}}} &
  {\color[HTML]{3166FF} \textbf{9.38}} \\ \cline{3-6} 
 &
   &
  Attack 3 &
  {\color[HTML]{3166FF} \textbf{9.39}} &
  \multicolumn{1}{l|}{{\color[HTML]{3166FF} \textbf{9.38}}} &
  {\color[HTML]{3166FF} \textbf{9.38}} \\ \cline{3-6} 
\multirow{-16}{*}{Proposed Method} &
  \multirow{-4}{*}{Attack 4} &
  Attack 4 &
  {\color[HTML]{3166FF} \textbf{6.91}} &
  \multicolumn{1}{l|}{{\color[HTML]{3166FF} \textbf{7.74}}} &
  {\color[HTML]{3166FF} \textbf{6.33}} \\ \hline
\end{tabular}
}
    \end{minipage} 

    \label{tab:WithoutBG}%
\end{table*}
%%%%%%%%%%%%%%%%%%%%%%

%%%----------------------------------------
\begin{figure*}[htp] 
    \centering
    \begin{subfigure}{8cm}\includegraphics[width=1\textwidth]{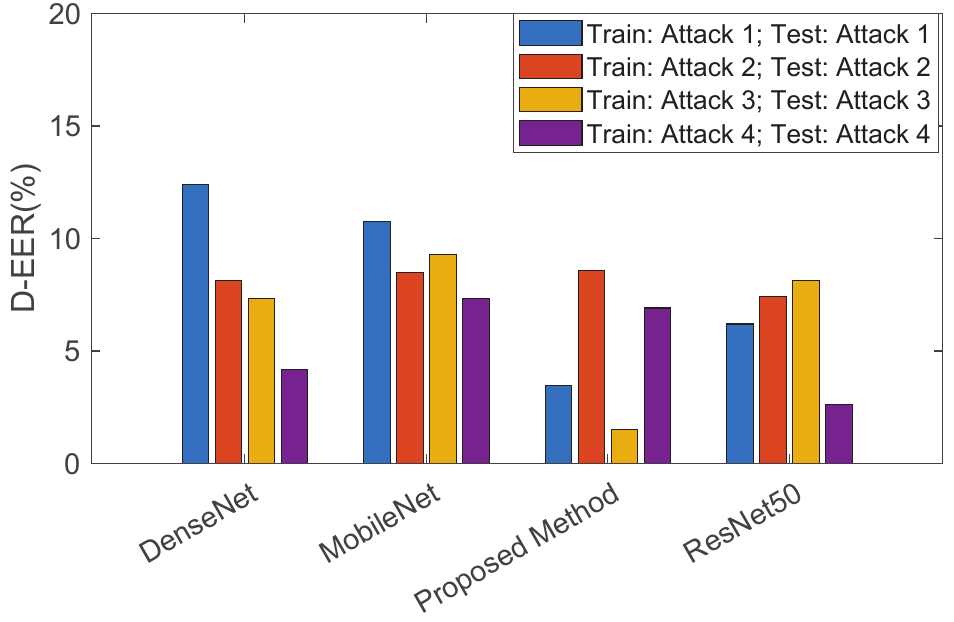}
    \caption{[D-EER(\%): Intra evaluation protocol}
    \label{fig:NoBGa1}
    \end{subfigure}

    \begin{subfigure}{8cm}\includegraphics[width=1\textwidth]{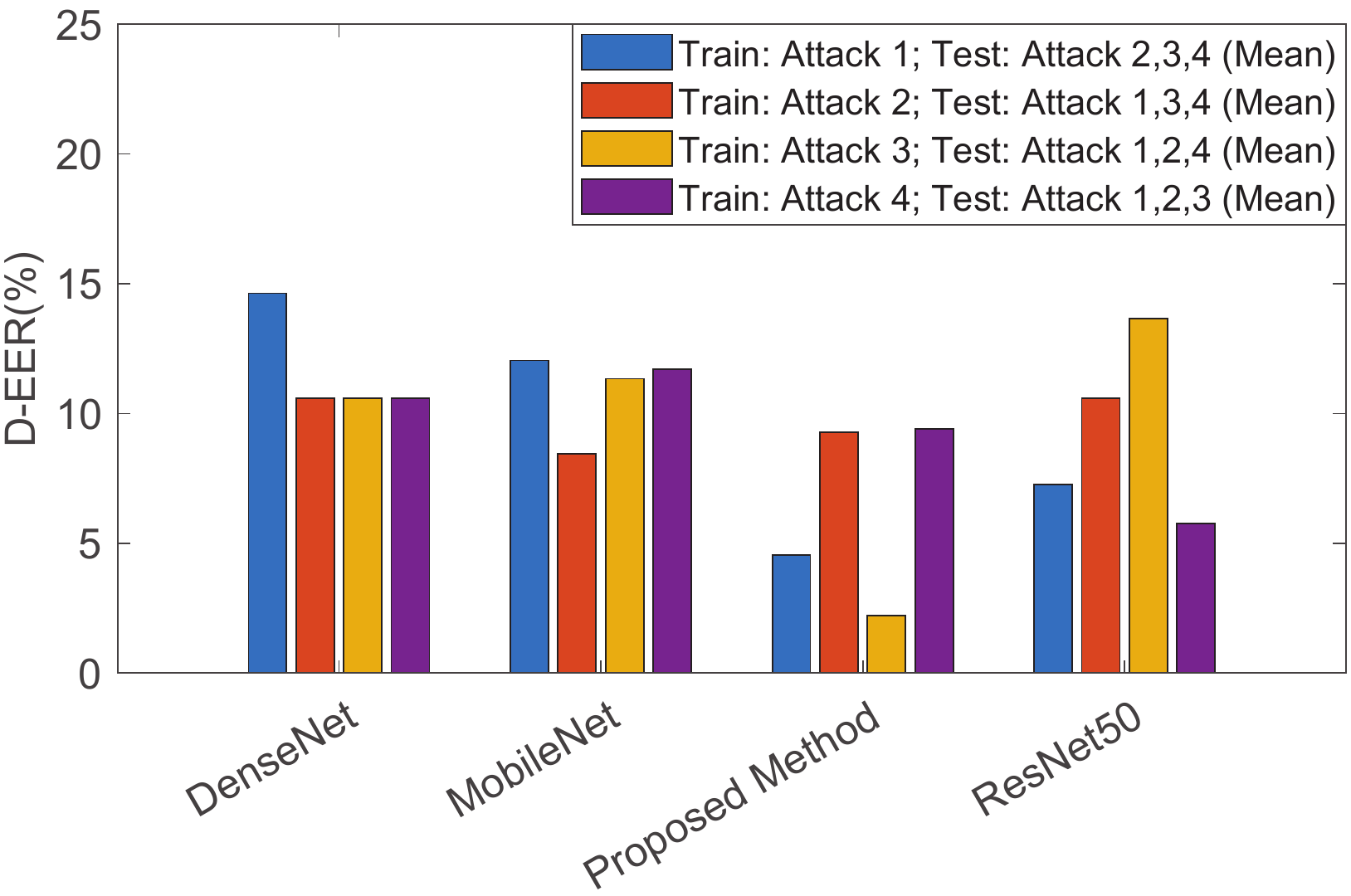}
    \caption{[D-EER(\%): Inter evaluation protocol}
    \label{fig:NoBGb2}
    \end{subfigure}
    \caption{D-EER(\%) of the proposed and existing methods on inter and intra experiments  \textbf{without background subtraction}}
\end{figure*}

%%%%----------------------------
\subsection{Results and Discussion: Without Background Subtraction}
This section discusses the quantitative performances of the proposed and existing PAD methods without background subtraction. Thus, the features were computed directly on the received signal, and experiments were performed using inter and intra evaluation protocols.   Table \ref{tab:WithoutBG} shows the quantitative performances of the proposed and existing PAD methods. Attack 1, indicated in Table \ref{tab:WithoutBG} corresponds to a display attack, Attack 2 corresponds to a print-I attack, Attack 3 corresponds to a print-II attack, and Attack 4 corresponds to a silicone face mask attack. Table \ref{tab:WithoutBG}  presents the quantitative results of both the intra and inter evaluation protocol and Figure \ref{fig:NoBGa1} and \ref{fig:NoBGb2} shows the bar chart with D-EER(\%) values for both intra and inter evaluation protocol. The bar chart in Figure \ref{fig:NoBGb2} indicates the inter evaluation protocol in which the D-EER (\%) is plotted by taking the mean of D-EER computed on test attacks. Based on the obtained results, the following important observations were made. 
\begin{itemize}
\item In general, the intra experiments indicate better results than the inter experiments on all four different PAIs. However, it is interesting to note that the difference in  performance between intra and inter experiments on PAIs is not much different, indicating the generalizability of the proposed acoustic signal analysis.  
\item Among the four PAIs employed in this work, the attack potential of these PAIs depends on the type of feature extraction. For example, Attack 1 indicates the highest D-EER (\%) with DenseNet features and Attack 3 indicates the highest D-EER (\%) with MobileNet and ResNet50. Attack 2 indicated the highest D-EER of the proposed method (\%).  
\item The proposed feature extraction using Efficientnet has indicated the best performance on Attacks 1 and 3 in inter and intra-experiments compared to the three different pre-trained networks employed in this work. However, the proposed method indicated less performance variation between  intra and inter evaluation protocols.  
\end{itemize}

%%%%%%%%%%%%%%%%%% with back ground Subtraction %%%%%%%%
\begin{table*}[!htb]
    \caption{Quantitative performance of the propose method with existing pre-trained \textbf{with back ground subtraction}}
    \begin{minipage}{.5\linewidth}
      %\caption{}
\centering
\resizebox{.9\columnwidth}{!}{%
\begin{tabular}{|l|l|l|l|ll|}
\hline
\multirow{2}{*}{Algorithms} & \multirow{2}{*}{Train Data} & \multirow{2}{*}{Test Data} & \multirow{2}{*}{D-EER} & \multicolumn{2}{l|}{BPCER @ APCER =} \\ \cline{5-6} 
 &                           &          &       & \multicolumn{1}{l|}{5\%}   & 10\%  \\ \hline
\multirow{16}{*}{DenseNet}  & \multirow{4}{*}{Attack 1}   & Attack 1                   & 9.41                   & \multicolumn{1}{l|}{15.96}   & 7.74  \\ \cline{3-6} 
 &                           & Attack 2 & 15.4  & \multicolumn{1}{l|}{21.59} & 14.31 \\ \cline{3-6} 
 &                           & Attack 3 & 13.51 & \multicolumn{1}{l|}{31.22} & 12    \\ \cline{3-6} 
 &                           & Attack 4 & 8.19  & \multicolumn{1}{l|}{11.73} & 5.86  \\ \cline{2-6} 
 & \multirow{4}{*}{Attack 2} & Attack 1 & 11.77 & \multicolumn{1}{l|}{31.22} & 15.13 \\ \cline{3-6} 
 &                           & Attack 2 & 6.07  & \multicolumn{1}{l|}{11.13} & 2.34  \\ \cline{3-6} 
 &                           & Attack 3 & 20.27 & \multicolumn{1}{l|}{58.45} & 37.79 \\ \cline{3-6} 
 &                           & Attack 4 & 10.59 & \multicolumn{1}{l|}{23.47} & 11.26 \\ \cline{2-6} 
 & \multirow{4}{*}{Attack 3} & Attack 1 & 6.81  & \multicolumn{1}{l|}{7.98}  & 4.69  \\ \cline{3-6} 
 &                           & Attack 2 & 4.12  & \multicolumn{1}{l|}{4.22}  & 2.81  \\ \cline{3-6} 
 &                           & Attack 3 & 5.38  & \multicolumn{1}{l|}{6.8}   & 3.23  \\ \cline{3-6} 
 &                           & Attack 4 & 4.17  & \multicolumn{1}{l|}{3.52}  & 2.34  \\ \cline{2-6} 
 & \multirow{4}{*}{Attack 4} & Attack 1 & 7.06  & \multicolumn{1}{l|}{11.73} & 4.69  \\ \cline{3-6} 
 &                           & Attack 2 & 7.44  & \multicolumn{1}{l|}{10.56} & 4.69  \\ \cline{3-6} 
 &                           & Attack 3 & 14.65 & \multicolumn{1}{l|}{44.36} & 26.29 \\ \cline{3-6} 
 &                           & Attack 4 & 4.17  & \multicolumn{1}{l|}{3.28}  & 2.11  \\ \hline
\end{tabular}%
}

%\caption{}
    \end{minipage}%
    \begin{minipage}{.5\linewidth}
      \centering
\resizebox{.9\columnwidth}{!}{%
\begin{tabular}{|l|l|l|l|ll|}
\hline
\multirow{2}{*}{Algorithms} & \multirow{2}{*}{Train Data} & \multirow{2}{*}{Test Data} & \multirow{2}{*}{D-EER} & \multicolumn{2}{l|}{BPCER @ APCER =} \\ \cline{5-6} 
 &                           &          &       & \multicolumn{1}{l|}{5\%}   & 10\%  \\ \hline
\multirow{16}{*}{ResNet50}  & \multirow{4}{*}{Attack 1}   & Attack 1                   & 3.46                   & \multicolumn{1}{l|}{3.05}   & 2.11   \\ \cline{3-6} 
 &                           & Attack 2 & 6.64  & \multicolumn{1}{l|}{7.74}  & 5.39  \\ \cline{3-6} 
 &                           & Attack 3 & 6.75  & \multicolumn{1}{l|}{7.98}  & 4.46  \\ \cline{3-6} 
 &                           & Attack 4 & 3     & \multicolumn{1}{l|}{2.58}  & 2.11  \\ \cline{2-6} 
 & \multirow{4}{*}{Attack 2} & Attack 1 & 7.56  & \multicolumn{1}{l|}{14.78} & 5.88  \\ \cline{3-6} 
 &                           & Attack 2 & 3.32  & \multicolumn{1}{l|}{2.34}  & 2.11  \\ \cline{3-6} 
 &                           & Attack 3 & 20.16 & \multicolumn{1}{l|}{54.22} & 40.61 \\ \cline{3-6} 
 &                           & Attack 4 & 7.16  & \multicolumn{1}{l|}{10.19} & 3.99  \\ \cline{2-6} 
 & \multirow{4}{*}{Attack 3} & Attack 1 & 9.41  & \multicolumn{1}{l|}{13.38} & 9.38  \\ \cline{3-6} 
 &                           & Attack 2 & 9.39  & \multicolumn{1}{l|}{13.61} & 8.68  \\ \cline{3-6} 
 &                           & Attack 3 & 8.74  & \multicolumn{1}{l|}{13.84} & 7.98  \\ \cline{3-6} 
 &                           & Attack 4 & 5.88  & \multicolumn{1}{l|}{6.14}  & 3.52  \\ \cline{2-6} 
 & \multirow{4}{*}{Attack 4} & Attack 1 & 9.41  & \multicolumn{1}{l|}{15.49} & 8.65  \\ \cline{3-6} 
 &                           & Attack 2 & 12.13 & \multicolumn{1}{l|}{22.55} & 12.91 \\ \cline{3-6} 
 &                           & Attack 3 & 14.77 & \multicolumn{1}{l|}{37.79} & 20.65 \\ \cline{3-6} 
 &                           & Attack 4 & 3.26  & \multicolumn{1}{l|}{1.87}  & 1.17  \\ \hline
\end{tabular}%
}
%\caption{}
    \end{minipage}

 \bigskip
 
    \begin{minipage}{.5\linewidth}
      %\caption{}
    \centering
\centering
\resizebox{.9\columnwidth}{!}{%
\begin{tabular}{|l|l|l|l|ll|}
\hline
\multirow{2}{*}{Algorithms}   & \multirow{2}{*}{Train Data} & \multirow{2}{*}{Test Data} & \multirow{2}{*}{D-EER} & \multicolumn{2}{l|}{BPCER @ APCER =} \\ \cline{5-6} 
 &                           &          &       & \multicolumn{1}{l|}{5\%}   & 10\%  \\ \hline
\multirow{16}{*}{MobileNetV2} & \multirow{4}{*}{Attack 1}   & Attack 1                   & 9.66                   & \multicolumn{1}{l|}{15.25}   & 9.62  \\ \cline{3-6} 
 &                           & Attack 2 & 12.13 & \multicolumn{1}{l|}{25.58} & 14.78 \\ \cline{3-6} 
 &                           & Attack 3 & 10.76 & \multicolumn{1}{l|}{19.24} & 11.26 \\ \cline{3-6} 
 &                           & Attack 4 & 9.14  & \multicolumn{1}{l|}{11.97} & 6.57  \\ \cline{2-6} 
 & \multirow{4}{*}{Attack 2} & Attack 1 & 12.13 & \multicolumn{1}{l|}{25.82} & 13.84 \\ \cline{3-6} 
 &                           & Attack 2 & 8.13  & \multicolumn{1}{l|}{13.84} & 4.92  \\ \cline{3-6} 
 &                           & Attack 3 & 20.16 & \multicolumn{1}{l|}{50}    & 37.79 \\ \cline{3-6} 
 &                           & Attack 4 & 12.14 & \multicolumn{1}{l|}{19.71} & 13.34 \\ \cline{2-6} 
 & \multirow{4}{*}{Attack 3} & Attack 1 & 11.27 & \multicolumn{1}{l|}{23.23} & 12.67 \\ \cline{3-6} 
 &                           & Attack 2 & 8.7   & \multicolumn{1}{l|}{16.9}  & 6.87  \\ \cline{3-6} 
 &                           & Attack 3 & 11.33 & \multicolumn{1}{l|}{34.97} & 12.91 \\ \cline{3-6} 
 &                           & Attack 4 & 6.72  & \multicolumn{1}{l|}{7.74}  & 5.39  \\ \cline{2-6} 
 & \multirow{4}{*}{Attack 4} & Attack 1 & 8.42  & \multicolumn{1}{l|}{12.91} & 7.51  \\ \cline{3-6} 
 &                           & Attack 2 & 7.32  & \multicolumn{1}{l|}{10.32} & 5.39  \\ \cline{3-6} 
 &                           & Attack 3 & 9.39  & \multicolumn{1}{l|}{18.17} & 7.74  \\ \cline{3-6} 
 &                           & Attack 4 & 4.16  & \multicolumn{1}{l|}{3.51}  & 2.11  \\ \hline
\end{tabular}%
}

    \end{minipage}%
    \begin{minipage}{.5\linewidth}
 \centering
\resizebox{0.9\columnwidth}{!}{%
\begin{tabular}{|l|l|l|l|ll|}
\hline
 &
   &
   &
   &
  \multicolumn{2}{l|}{BPCER @ APCER =} \\ \cline{5-6} 
\multirow{-2}{*}{Algorithms} &
  \multirow{-2}{*}{Train Data} &
  \multirow{-2}{*}{Test Data} &
  \multirow{-2}{*}{D-EER} &
  \multicolumn{1}{l|}{5\%} &
  10\% \\ \hline
 &
   &
  Attack 1 &
  {\color[HTML]{3531FF} \textbf{1.85}} &
  \multicolumn{1}{l|}{{\color[HTML]{3531FF} \textbf{1.17}}} &
  {\color[HTML]{3531FF} \textbf{0.99}} \\ \cline{3-6} 
 &
   &
  Attack 2 &
  {\color[HTML]{3531FF} \textbf{2.06}} &
  \multicolumn{1}{l|}{{\color[HTML]{3531FF} \textbf{1.64}}} &
  {\color[HTML]{3531FF} \textbf{1.17}} \\ \cline{3-6} 
 &
   &
  Attack 3 &
  {\color[HTML]{3531FF} \textbf{1.94}} &
  \multicolumn{1}{l|}{{\color[HTML]{3531FF} \textbf{1.64}}} &
  {\color[HTML]{3531FF} \textbf{1.17}} \\ \cline{3-6} 
 &
  \multirow{-4}{*}{Attack 1} &
  Attack 4 &
  {\color[HTML]{3531FF} \textbf{1.17}} &
  \multicolumn{1}{l|}{{\color[HTML]{3531FF} \textbf{0.93}}} &
  {\color[HTML]{3531FF} \textbf{0.93}} \\ \cline{2-6} 
 &
   &
  Attack 1 &
  {\color[HTML]{3531FF} \textbf{7.92}} &
  \multicolumn{1}{l|}{{\color[HTML]{3531FF} \textbf{8.92}}} &
  {\color[HTML]{3531FF} \textbf{6.8}} \\ \cline{3-6} 
 &
   &
  Attack 2 &
  {\color[HTML]{3531FF} \textbf{5.38}} &
  \multicolumn{1}{l|}{{\color[HTML]{3531FF} \textbf{7.98}}} &
  {\color[HTML]{3531FF} \textbf{3.75}} \\ \cline{3-6} 
 &
   &
  Attack 3 &
  {\color[HTML]{3531FF} \textbf{9.39}} &
  \multicolumn{1}{l|}{{\color[HTML]{3531FF} \textbf{9.85}}} &
  {\color[HTML]{3531FF} \textbf{9.38}} \\ \cline{3-6} 
 &
  \multirow{-4}{*}{Attack 2} &
  Attack 4 &
  {\color[HTML]{3531FF} \textbf{5.61}} &
  \multicolumn{1}{l|}{{\color[HTML]{3531FF} \textbf{5.86}}} &
  {\color[HTML]{3531FF} \textbf{3.75}} \\ \cline{2-6} 
 &
   &
  Attack 1 &
  {\color[HTML]{3531FF} \textbf{2.85}} &
  \multicolumn{1}{l|}{{\color[HTML]{3531FF} \textbf{2.34}}} &
  {\color[HTML]{3531FF} \textbf{1.64}} \\ \cline{3-6} 
 &
   &
  Attack 2 &
  {\color[HTML]{3531FF} \textbf{1.49}} &
  \multicolumn{1}{l|}{{\color[HTML]{3531FF} \textbf{1.48}}} &
  {\color[HTML]{3531FF} \textbf{0.94}} \\ \cline{3-6} 
 &
   &
  Attack 3 &
  {\color[HTML]{3531FF} \textbf{1.48}} &
  \multicolumn{1}{l|}{{\color[HTML]{3531FF} \textbf{1.4}}} &
  {\color[HTML]{3531FF} \textbf{0.93}} \\ \cline{3-6} 
 &
  \multirow{-4}{*}{Attack 3} &
  Attack 4 &
  {\color[HTML]{3531FF} \textbf{1.44}} &
  \multicolumn{1}{l|}{{\color[HTML]{3531FF} \textbf{0.93}}} &
  {\color[HTML]{3531FF} \textbf{0.49}} \\ \cline{2-6} 
 &
   &
  Attack 1 &
  {\color[HTML]{3531FF} \textbf{2.14}} &
  \multicolumn{1}{l|}{{\color[HTML]{3531FF} \textbf{1.46}}} &
  {\color[HTML]{3531FF} \textbf{0.72}} \\ \cline{3-6} 
 &
   &
  Attack 2 &
  {\color[HTML]{3531FF} \textbf{2.18}} &
  \multicolumn{1}{l|}{{\color[HTML]{3531FF} \textbf{2.11}}} &
  {\color[HTML]{3531FF} \textbf{1.41}} \\ \cline{3-6} 
 &
   &
  Attack 3 &
  {\color[HTML]{3531FF} \textbf{2.16}} &
  \multicolumn{1}{l|}{{\color[HTML]{3531FF} \textbf{2.11}}} &
  {\color[HTML]{3531FF} \textbf{0.99}} \\ \cline{3-6} 
\multirow{-16}{*}{Proposed Method} &
  \multirow{-4}{*}{Attack 4} &
  Attack 4 &
  {\color[HTML]{3531FF} \textbf{0.64}} &
  \multicolumn{1}{l|}{{\color[HTML]{3531FF} \textbf{0.23}}} &
  {\color[HTML]{3531FF} \textbf{0.23}} \\ \hline
\end{tabular}%
}

    \end{minipage} 
     \label{tab:WithBG}%
\end{table*}
%%%%%%%%%%%%%%%%%%%%%%

%%%----------------------------------------
\begin{figure*}[htp] 
    \centering
    \begin{subfigure}{8cm}
     \includegraphics[width=1\textwidth]{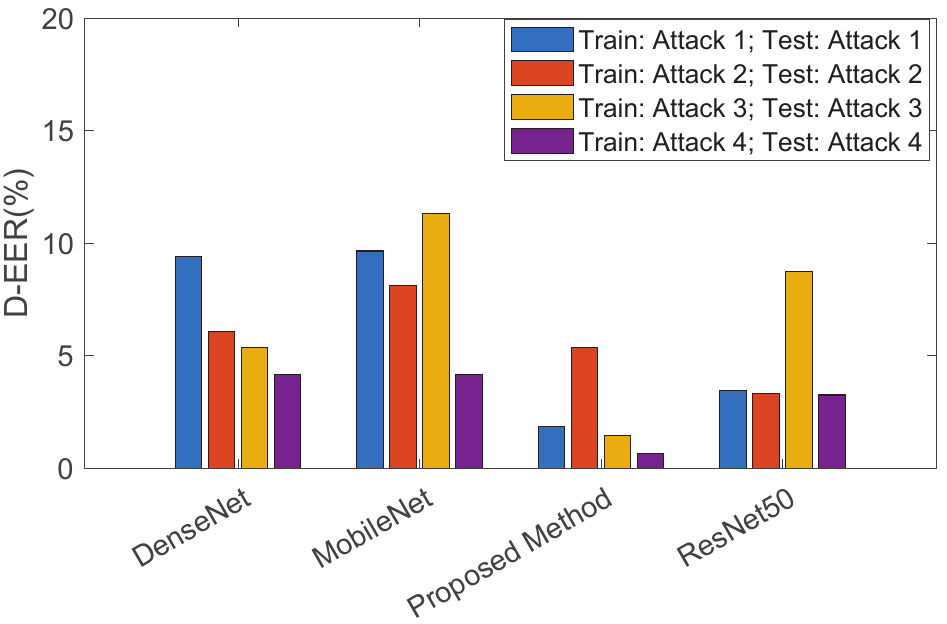}
     \caption{D-EER(\%): Intra evaluation protocol}
     \label{fig:BGa1}
    \end{subfigure}
 \begin{subfigure}{8cm}
     \includegraphics[width=1\textwidth]{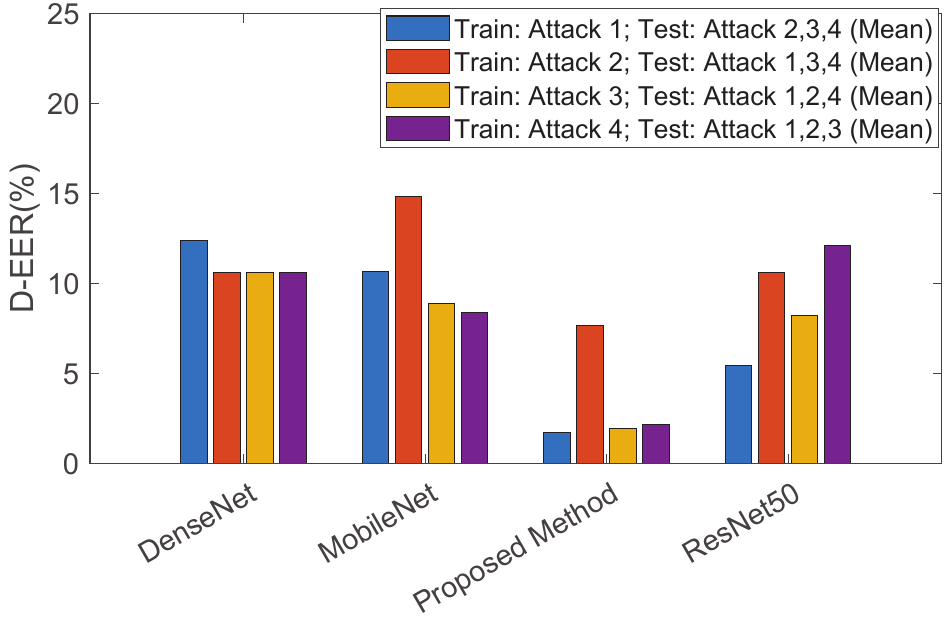}
     \caption{D-EER(\%): Inter evaluation protocol}
      \label{fig:BGa2}
    \end{subfigure}
    \caption{D-EER(\%) of the proposed and existing methods on inter and intra experiments  \textbf{with background subtraction}}
\end{figure*}

%%%%----------------------------

%%%----------------------------------------
\begin{figure*}[htp] 
    \centering
      \begin{subfigure}{8cm}
       \includegraphics[width=1\textwidth]{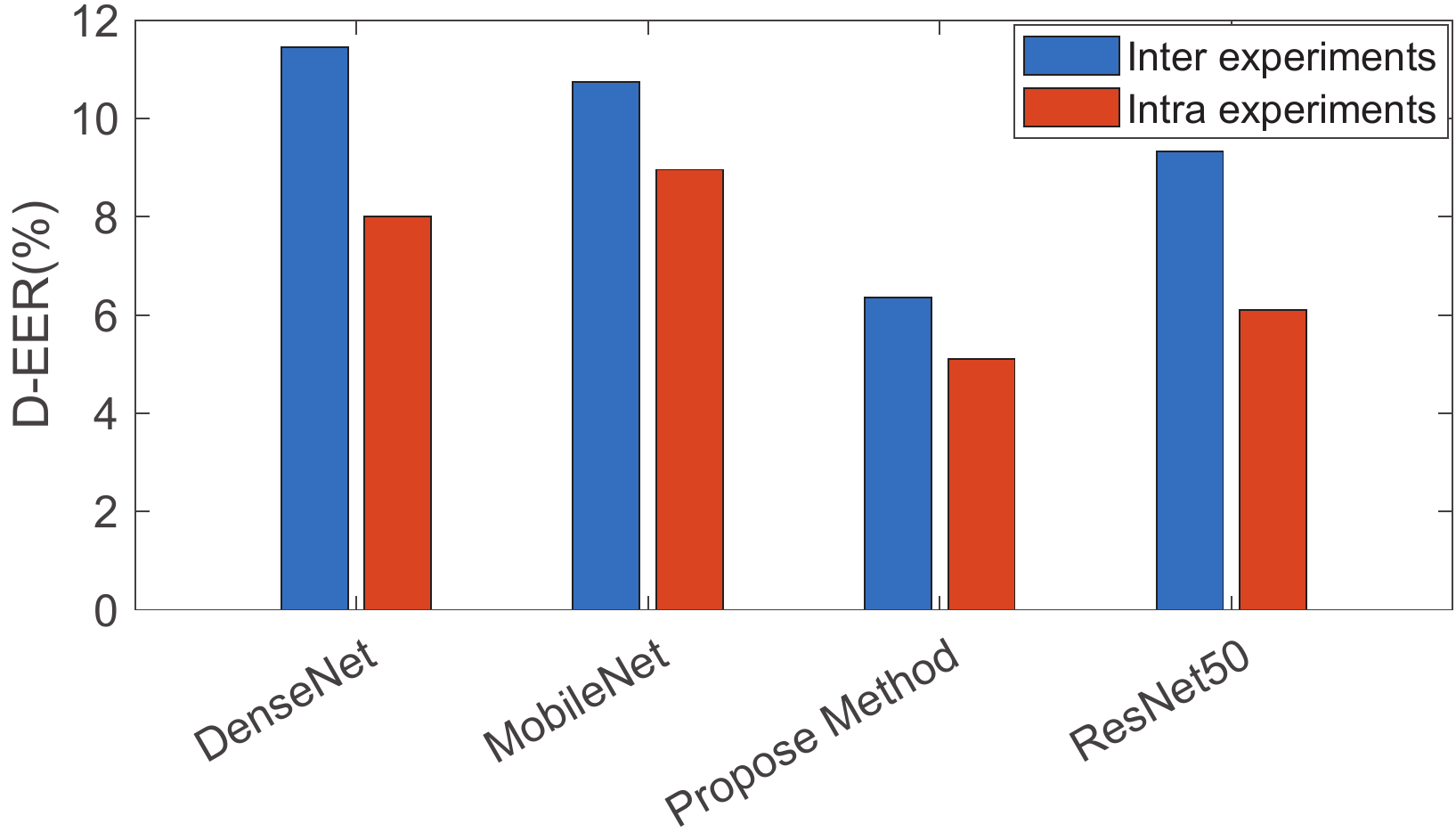}
       \caption{D-EER(\%) performance without background Subtraction}
          \label{fig:a}
      \end{subfigure}

      \begin{subfigure}{8cm}
       \includegraphics[width=1\textwidth]{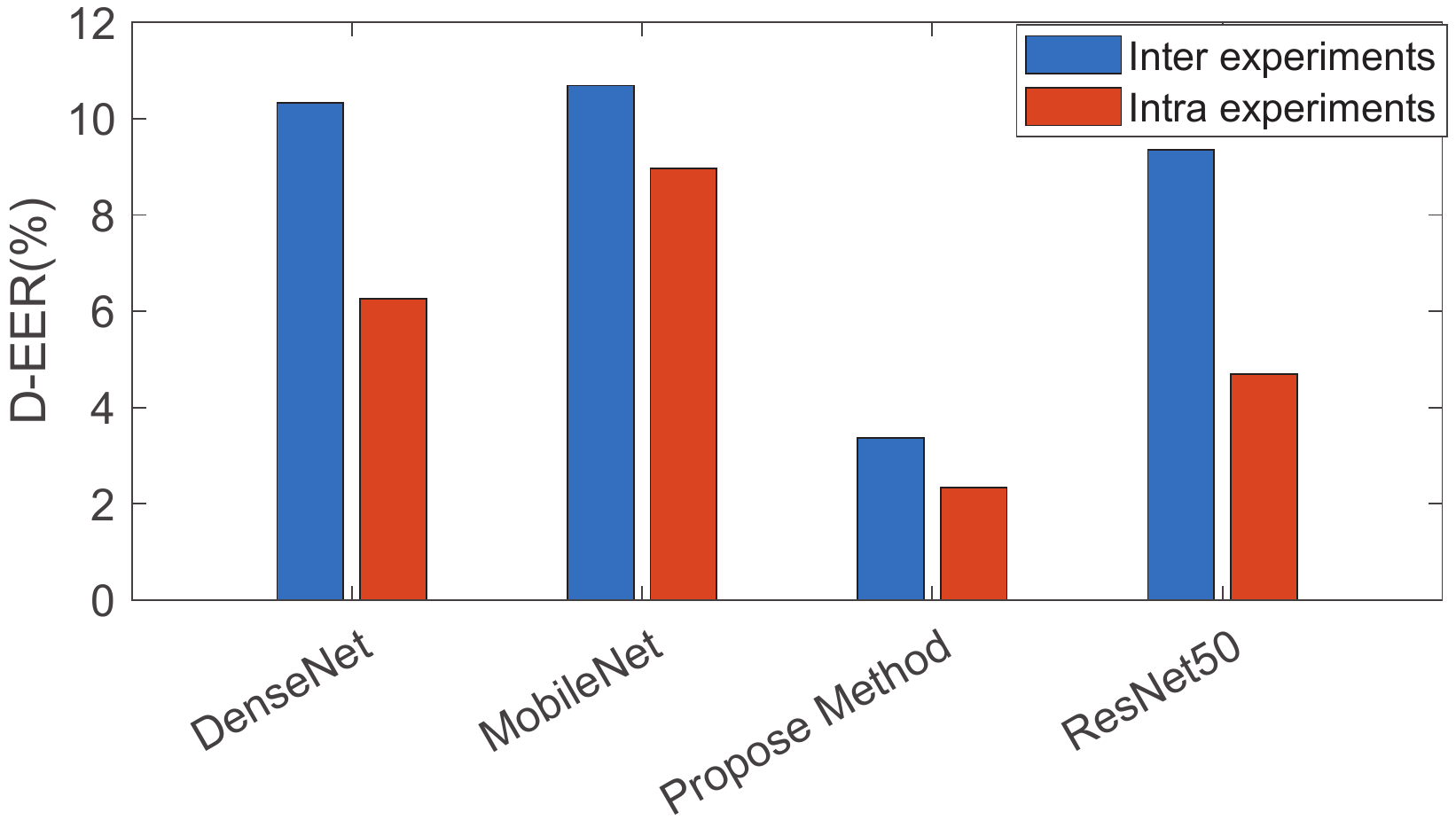}
       \caption{D-EER(\%) performance with background Subtraction}
          \label{fig:b}
      \end{subfigure}
    \caption{Average D-EER(\%) of inter and intra experiments with and without background subtraction}
\end{figure*}
\subsection{Results and Discussion: With Background Subtraction}
This section discusses the quantitative results of the proposed and existing PAD methods when background subtraction was performed. The uniqueness of the proposed transmission signal lies in its ability to record the background before the signal is transmitted and received. Therefore, we can subtract the background signal from the received signal to improve the SNR and contribute to reliable detection of PAI.   Table \ref{tab:WithBG} presents the quantitative performances of the proposed and existing PAD techniques with inter  and intra evaluation protocols. Figure \ref{fig:BGa1} and \ref{fig:BGa2} show bar charts with the D-EER(\%) for both intra  and inter experiments. Based on the results obtained, the following are important observations:

\begin{itemize}
\item The detection error is less for the intra experiments compared to the inter experimental protocol with both proposed and existing feature extraction techniques. However, the average difference in performance between the intra and inter protocols was minimal. Therefore, the use of acoustic signals can result in a generalizable PAD. 
\item Among the four PAIs employed in this work, the attack potential of these PAIs depends on the type of feature extraction. For example, Attack 1 indicates the highest D-EER (\%) with DenseNet features, and Attack 3 indicates the highest D-EER (\%) with MobileNet and ResNet50. Attack 2 indicates the proposed method's highest D-EER (\%). 
\item The proposed feature extraction using Efficientnet has indicated the best performance on inter and intra experiments compared to the three different pre-trained networks employed in this work. The results indicated the robustness of the proposed method to background noise, as background noise subtraction was performed in these experiments. 
\end{itemize}

%%%%----------------------------
Figure \ref{fig:a} and \ref{fig:b} show the average performance of the proposed and existing feature extraction methods in the inter and intra experiments with and without the background subtraction method. The following are the important observations:
\begin{itemize}
\item The detection performance of the PAD algorithms indicates improved performance when background subtraction is performed. This demonstrated the superiority of the proposed transmission signal. 
\item The proposed feature extraction based on the Efficientnet indicates the best performance on  with and without background subtraction compared to other feature extraction techniques.  
\item The proposed method indicates the little difference between intra and inter-performance variation on both with and without background subtraction. The proposed method indicates an average D-EER (\%) of 6.35(\%) and 5.11(\%) on inter and intra experiments without using background subtraction. With background subtraction, the proposed method indicated an average D-EER (\%) of 3.36(\%) and 2.33(\%) in inter  and intra experiments respectively. The low difference in the performance of the proposed method with inter and intra variations indicates the generalizability of the proposed method. 
\end{itemize}

%------------------------------------------------------------------------
\section{Conclusion and Future Work}
\label{sec:Conc}

Reliable detection of unknown PA is essential for enabling trustworthy face recognition applications on smartphones. In this study, we presented a novel method for a generalizable face PAD on smartphones using acoustic sound echoes. A novel signal based on the wide pulse is proposed to effectively model the background noise and increase the signal-to-noise ratio. The reflected signals were processed to remove background noise and obtain the time-frequency representation. We then computed the deep features using pre-trained EfficientNet by extracting the features from the BatchNorm layer. The BatchNorm layer provides 49 different embeddings used to train 49 independent linear SVMs whose decisions are fused to make the final decision. Extensive experiments are presented to benchmark the performance of the proposed method using intra and inter evaluation protocols. Additional experiments are presented to highlight the importance of background subtraction in improving the robustness and accuracy of the face PAD. The obtained results demonstrated the generalizability of the proposed method across unknown PAIs. Future work will extend the proposed method  to different types of smartphones. Further extensive data collection is planned in the noisy conditions to benchmark the PAD algorithms based on acoustic reflections. 

%------------------------------------------------------------------------
{\small
\bibliographystyle{ieee}
\bibliography{main}
}

\end{document}